\documentclass{article} 
\usepackage{iclr2019_conference,times}

\usepackage{amsmath,amsfonts,bm}

\def\eqref#1{equation~\ref{#1}}

\def\1{\bm{1}}

\DeclareMathAlphabet{\mathsfit}{\encodingdefault}{\sfdefault}{m}{sl}
\SetMathAlphabet{\mathsfit}{bold}{\encodingdefault}{\sfdefault}{bx}{n}

\usepackage{hyperref}
\usepackage{url}
\usepackage{graphicx}
\usepackage{textcomp}
\usepackage{caption}
\usepackage{tabulary}

\let\<\textless
\let\>\textgreater
\newcommand{\specialcell}[2][c]{%
  \begin{tabular}[#1]{@{}c@{}}#2\end{tabular}}

\usepackage{subcaption}
\usepackage{CJKutf8}
\usepackage{array}
\usepackage{enumitem}
\usepackage{amssymb}
\usepackage{xcolor}
\usepackage{etoolbox}

\newenvironment{sequation}{\begin{equation}}{\end{equation}}

\graphicspath{{./figures/}}

\title{Generalize Symbolic Knowledge With\\ Neural Rule Engine}

\makeatletter
\def\thanks#1{\protected@xdef\@thanks{\@thanks
        \protect\footnotetext{#1}}}
\makeatother

\author{
	Shen Li $^*$\thanks{*\:shen@deeplycurious.ai} \And
	Hengru Xu $^\dag$\thanks{\dag\:Work performed when he worked as an intern from Beijing University of Post and Telecommunication.}\And
	Zhengdong Lu \AND\\
	Deeplycurious.ai
}

\iclrfinalcopy 
\begin{document}
\bibliographystyle{apalike}

\maketitle

\begin{CJK}{UTF8}{gbsn}

\begin{abstract}
As neural networks have dominated the state-of-the-art results in a wide range of NLP tasks, it attracts considerable attention to improve the performance of neural models by integrating symbolic knowledge. Different from existing works, this paper investigates the combination of these two powerful paradigms from the knowledge-driven side. We propose Neural Rule Engine (NRE), which can learn knowledge explicitly from logic rules and then generalize them implicitly with neural networks. NRE is implemented with neural module networks in which each module represents an action of a logic rule. The experiments show that NRE could greatly improve the generalization abilities of logic rules with a significant increase in recall. Meanwhile, the precision is still maintained at a high level.
\end{abstract}

\section{Introduction}
Human cognition successfully integrates the connectionist and symbolic paradigms. 
Yet the modelling of cognition develops separately in neural computation and symbolic logic areas \citep{garcez2012neural}. 
Neural networks are well known for their inductive learning and generalization capabilities on large amounts of data, while symbolic logic rules are mostly constructed with expert knowledge, which yields high precision and good interpretabilities.

\cite{minsky1991logical} states that both symbolism and connectionist have virtues and deficiencies, and we need integrated systems that can exploit the advantages of both. Recently, there is a movement towards a fruitful combination of these two streams, e.g. integrating symbolic knowledge into neural networks \citep{hu2016harnessing,lu2017object,luo2018marrying}.
As pre-defined symbolic knowledge can greatly improve the learning effectiveness of neural models, it raises the question that can neural networks help to improve the generalization abilities of rules? 

Nowadays, symbolic knowledge is still widely used in few data scenarios or combined with statistical models. 
However, logic rules built with symbolic knowledge have poor generalization abilities and thus relatively low recalls. 
For human it is very natural to learn a piece of knowledge at first, and then use a couple of cases to generalize the knowledge. 
Inspired by this learning strategy, we propose a neural rule engine (NRE), in which rules can acquire higher flexibility and generalization ability with the help of neural networks, while they can still maintain the advantages of high precision and interpretabilities.

In this paper, we construct logic rules with the most widespread regular expressions (REs). NRE transforms the rules into neural module networks (NMN) which have symbolic structures. Specifically, the transformation involves 2 steps:
\begin{itemize}[leftmargin=*]
\begin{item}
Parse a RE into an action tree composed of finite pre-defined actions. This operation is inspired by \cite{kaplan1994regular}'s work where each RE is considered as a finite-state machine (FSM). The types and orders of actions are determined by a neural parser or a symbolic parser.
\end{item}
\begin{item}
Represent the RE actions as neural-symbolic modules. Each module can be either customized neural networks, or a symbolic algorithm.
\end{item}
\end{itemize}
With neural rule engine, a system can learn knowledge explicitly from logic rules and generalize them implicitly with neural networks. It is not only an innovative paradigm of neural-symbolic learning, but also an effective solution to real life applications, including improving the existing rule-based systems and building neural rule methods for applications which do not have sufficient training data.

\section{Related Work}
\textbf{Neural symbolic learning}
Neural-symbolic systems have been applied to various tasks, including ontology learning, 
fault diagnosis, robotics, training and assessment in simulators \citep{hitzler2005ontology,de2011neural,garcez2015neural}.
Recently, there are other efforts in the topical proximity of the core field in neural-symbolic integration \citep{besold2017neural}, including paradigms in computation and representation such as ``conceptors'' \citep{jaeger2014controlling}, ``Neural Turing Machines'' (NTMs) \citep{graves2014neural}.

In addition, a line of research aims to encode symbolic rules or prior knowledge into neural networks.
\cite{hu2016harnessing} harness neural networks by logic rules, transferring the structured information into the network parameters.
\cite{xiao2017symbolic} exploit prior knowledge such as weighted context-free grammar and the likelihood that entities occur in the input to form the ``background'' to RNN models.
\cite{lu2017object} present Object-oriented Neural Programming, allowing neural and symbolic representing and reasoning over complex structures for semantic parsing.
\cite{li2017initializing} encode semantic features into CNNs filters instead of initializing them randomly.
\cite{wang2017combining} propose a framework based on CNN where text representations are incorporated with words and relevant concepts conceptualized by rules in a knowledge base.  
\cite{luo2018marrying} exploit expressiveness of regular expressions at different levels of a neural network, aiming to use the information provided by rules to improve the performance of the network.
Different from existing works which enhance the performance of NNs by introducing prior knowledge, this work focuses on the knowledge side, and aims to improve logic rules with NNs.

\textbf{Neural module networks}
Neural Module Networks (NMN), first proposed in Visual Question Answering (VQA), are composed of collections of joint-trained neural models.

\cite{andreas2016neural} use a rule-based dependency parser to determine the basic computational modules needed to answer the question, as well as the relations between the modules.
\cite{andreas2016learning} present a model for selecting layouts from a set of structures automatically generated by a dependency parser, which is called Dynamic Neural Module Network.
\cite{hu2017learning} propose End-to-End Module Networks, which learn to generate network structures with a neural parser.

Since each RE can be considered as a finite-state machine (FSM) \citep{kaplan1994regular}, we can use finite pre-defined actions to interpret REs.
Inspired by these works, we propose Neural Rule Engine (NRE), where REs are composed of finite, reusable, computational and joint-trained modules.
With NRE, knowledge is learned explicitly from logic rules and generalized implicitly with neural networks.

\section{Neural Rule Engine}

It is worth noting that the learning process of human does not rely on only the data (cases) or the logic knowledge, but the combination of them. For example, when a father introduces the concept of bird to his child, he may point to a visual bird (a live one or an image), and tell his child this flying animal with feathers is a bird. Given the knowledge of bird and a couple of visual cases, the child becomes capable of recognizing a bird.
Inspired by this, we propose a novel strategy to teach models to learn rules like children, where we impart a rule to models at first, and then use a couple of cases to help models generalize the knowledge.

\subsection{Architecture}

\begin{figure}[h]
\begin{center}
\includegraphics[width=0.7\linewidth]{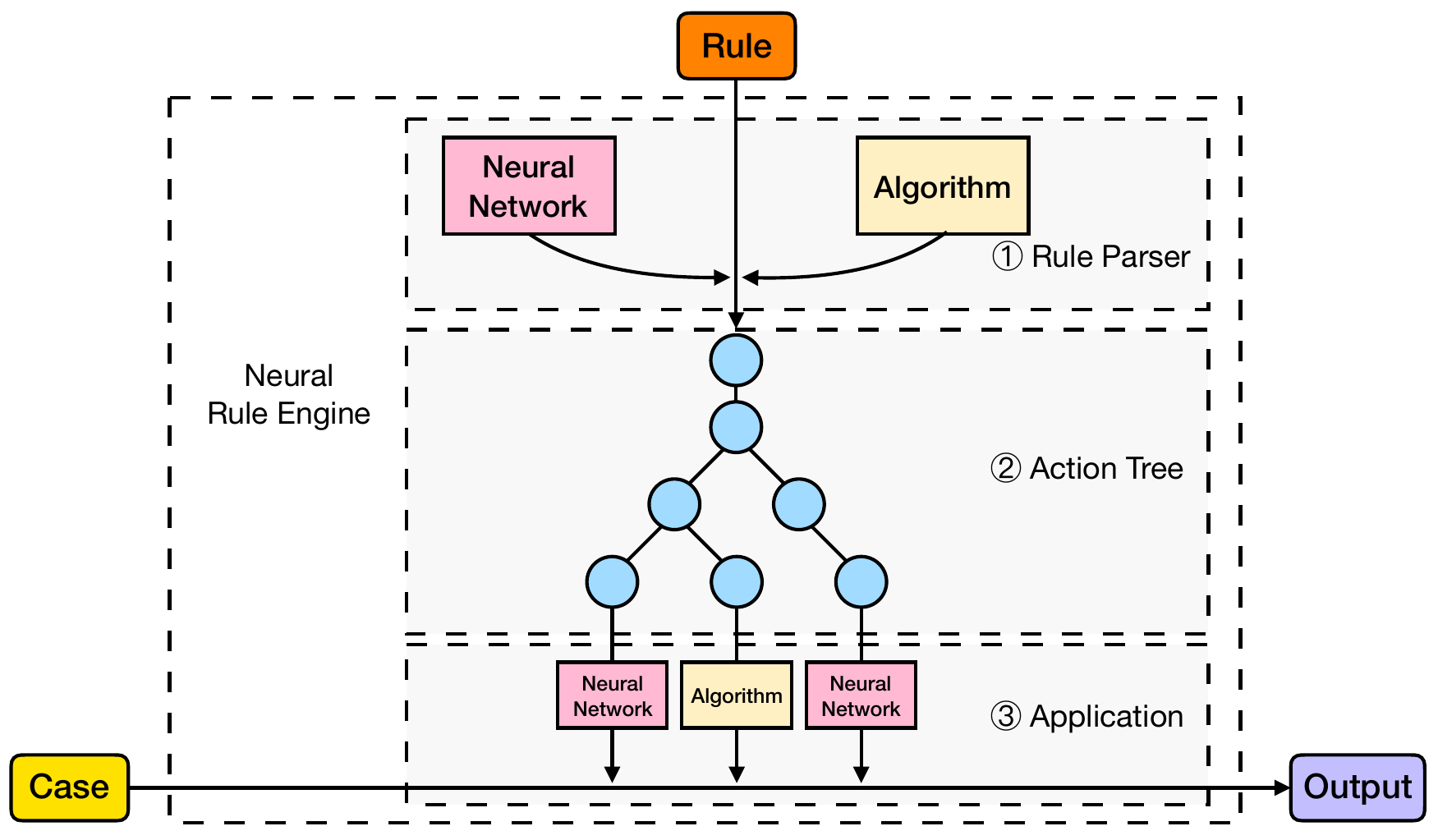}
\end{center}
\caption{
NRE includes action modules and a rule parser, both of which can either be trained neural networks or predefined algorithms.
The circles in ``Action Tree'' are actions with corresponding parameters predicted by the ``Rule Parser''.
}
\label{fig: model_overview}
\end{figure}

\begin{figure}[h]
\begin{center}
\includegraphics[width=0.7\linewidth]{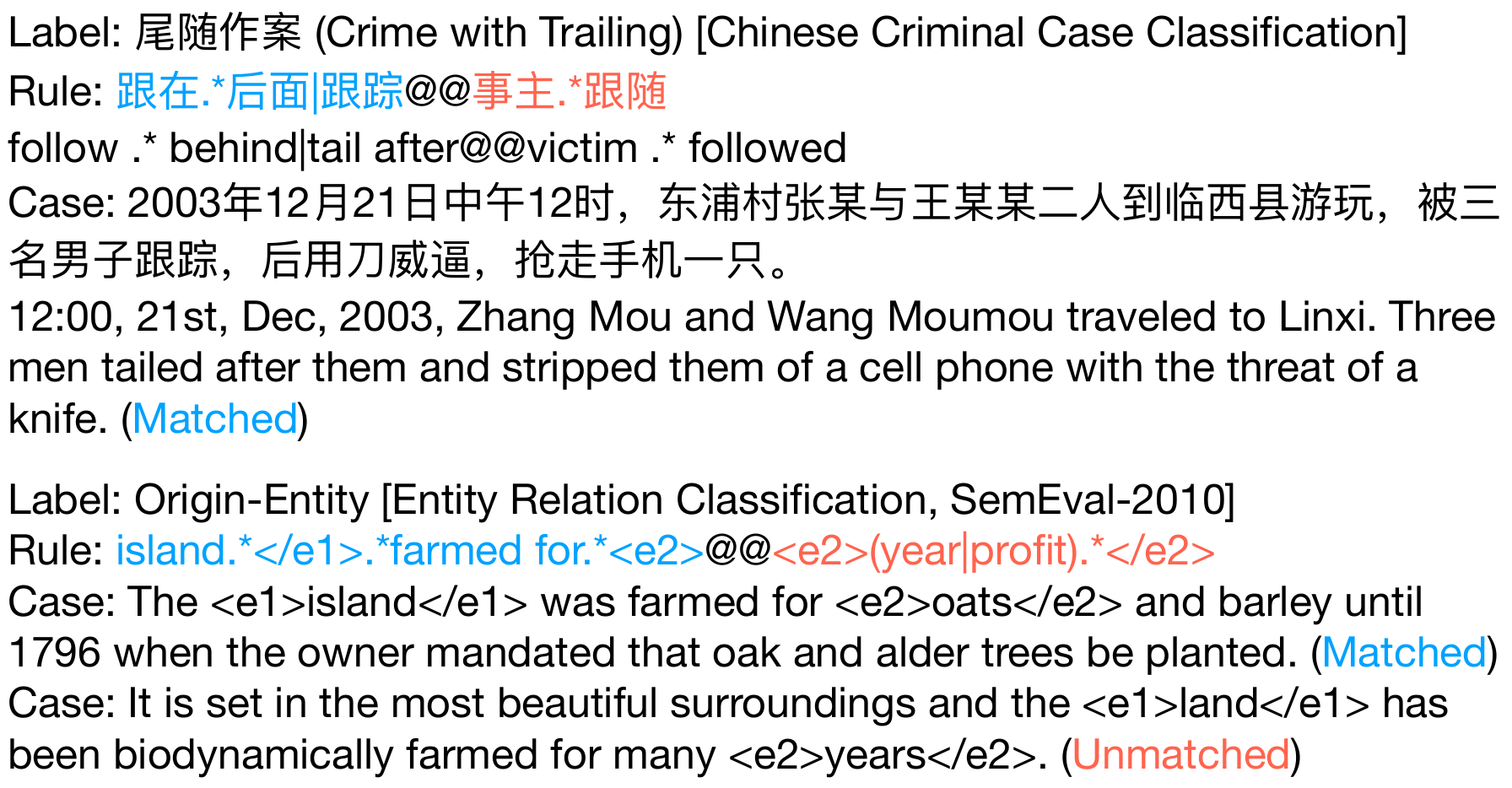}
\end{center}
\caption{Rule examples.}
\label{fig: rule example}
\end{figure}

NRE consists of two main components: a set of actions (neural modules or simply mathematical functions) that provide basic operations, and a layout parser (neural networks or predefined algorithms) to predict a specific layout for every RE. With this layout, modules are dynamically assembled. 
An overview pipeline of our model is shown in Figure \ref{fig: model_overview}.
NRE is interpretable and flexible, which can be considered as an ``enhanced'' rule engine.

NRE can be widely applied to a variety of tasks.
The output can be a labeled sequence, whose length is the same as the case. 
In this situation, NRE can be served for sequence labeling task.
Besides, NRE can determine whether a case is matched with a rule and output a label for classification task. 

In this paper, we use REs to represent rules. 
Figure \ref{fig: rule example} illustrates two classification rule examples. 
We define a rule as: $r_{y_{m} i} = [r_{0}, \cdots, r_{k}]$ and a case as: $x = [x_{0}, \cdots, x_{n}]$ where $r_{y_{m} i}$ denotes the $i$th rule corresponding to label $y_{m}$, $r_{k}$ is a token in the regular expression, and $x_{n}$ is a token in the case.
If rule $r_{y_{m} i}$ matches case $x$, $r_{y_{m} i}(x) = 1$, and if not, $r_{y_{m} i}(x) = 0$.
Given a case $x$, the model needs to choose the correct labels from $y = [y_{0}, \cdots, y_{m}]$.

In our method, each rule has a positive part and a negative part, which are separated by ``@@''.
The positive part is what should appear in a case and the negative shouldn't.
As shown in the second case in Figure \ref{fig: rule example}, rule $r$ matches case $x$ when $island.*\langle/e1\rangle.*farmed \: for.*\langle e2 \rangle$ matches $x$ and $\langle e2 \rangle(year|profit).*\langle /e2 \rangle$ doesn't.
In this example, the relation between $e1$ and $e2$ will be labeled as ``Origin-Entity'' if the rule matches the case.

\begin{figure}[h]
\begin{center}
\includegraphics[width=0.7\linewidth]{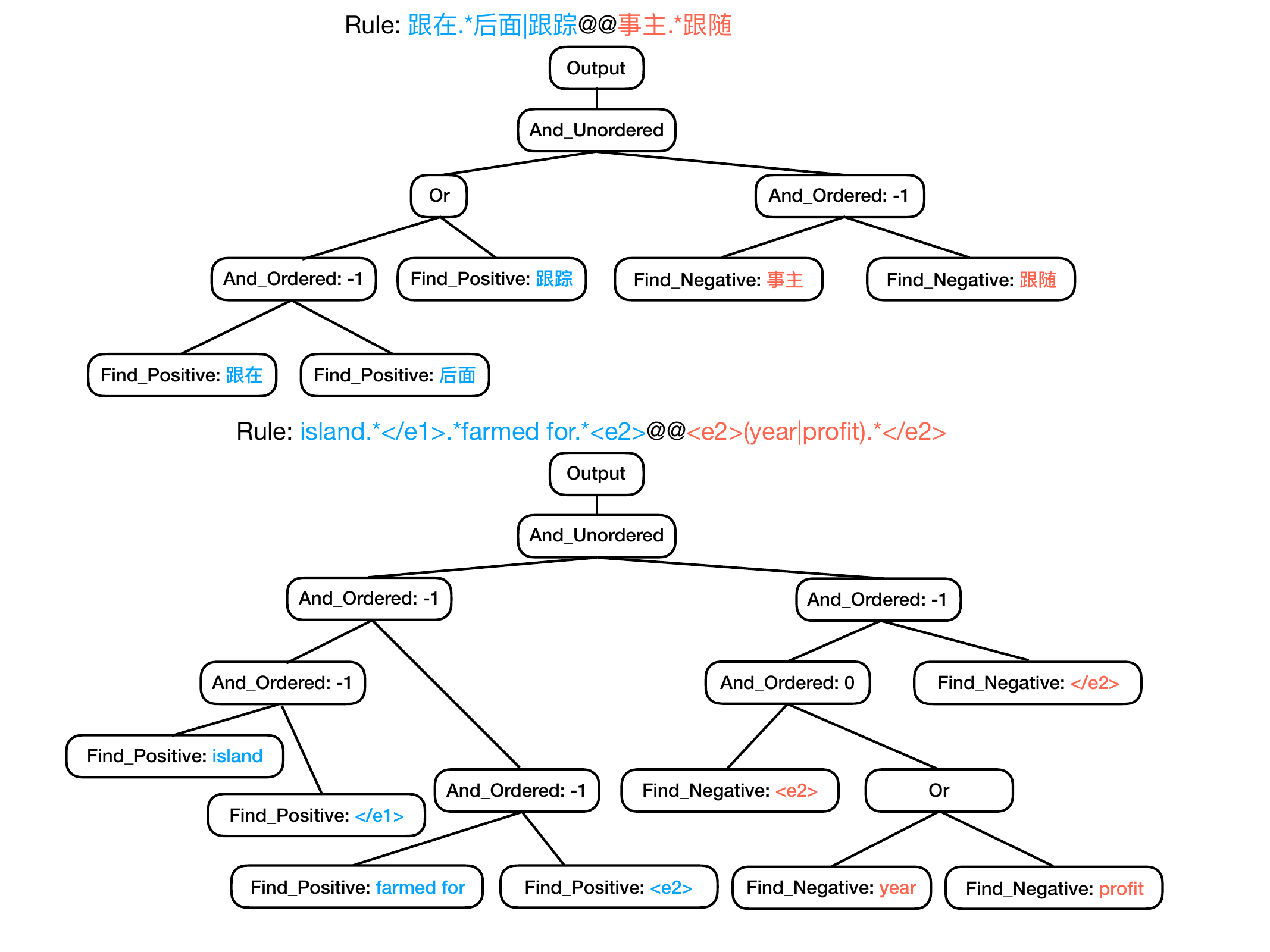}
\end{center}
\caption{A rule is parsed to a tree structure. Each node is an action with a specific parameter and its child nodes which are served as the inputs.}
\label{fig: rule_disassemble}
\end{figure}

\begin{figure}[h]
\begin{center}
\includegraphics[width=0.7\linewidth]{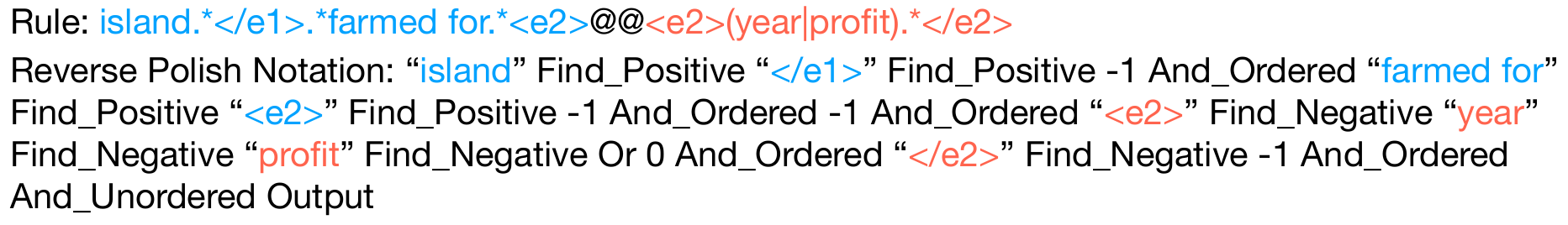}
\end{center}
\caption{An example showing how to linearize a RE to a sequence of modules and their parameters.}
\label{fig: rpn_example}
\end{figure}

Given a rule $r_{y_{m} i} = [r_{0}, \cdots, r_{k}]$, the parser firstly predicts a functional expression $f_{y_{m} i} = [a_{0}(p_{0}), a_{1}(p_{1}), \cdots, a_{j}(p_{j})]$, listed in ``Action Tree'' (Figure \ref{fig: rule_disassemble}), consisting of actions and parameters, where $a_{j}$ indicates the action and $p_{j}$ indicates the parameter for action $a_{j}$.
In implementation, we use Reverse Polish Notation (RPN) \citep{burks1954analysis}, post-order traversal over the syntax tree, to represent REs (Figure \ref{fig: rpn_example}).
A network is assembled with the modules according to the specific RPN to get the final output.

\subsection{Neural Modules}
\label{sec: Neural Modules}

\begin{table}[!htb]
\begin{center}
\begin{tabular}{cc}
\hline
\multicolumn{1}{c}{\bf Action}  &\multicolumn{1}{c}{\bf Detail} \\
\hline
\hline
\specialcell{\boldmath{$Find\_Positive(x)$} \\Find tokens matching $x$.} &NN / $re.match(x)$ \\\hline
\specialcell{\boldmath{$Find\_Negative(x)$} \\Find tokens matching $x$.} &NN / $re.match(x)$ \\\hline
\specialcell{\boldmath{$And\_Ordered(x_1, x_2, d)$} \\Determine if the distance between \\$x_1$ and $x_2$ meets $d$.}  &NN / $x_1\{, d\}x_2$ \\\hline
\specialcell{\boldmath{$And\_Unordered(x_1, x_2)$} \\Determine if $x_1$ and $x_2$ co-occur.} &$(x_1 + x_2) * sum(x_1) * sum(x_2)$\\\hline
\specialcell{\boldmath{$Or(x_1, x_2)$} \\Determine if $x_1$ or $x_2$ occurs.} &$x_1 + x_2$\\\hline
\specialcell{\boldmath{$Output(x)$} \\Output labels.} &$max(x)$\\\hline
\end{tabular}
\end{center}
\caption{Actions}
\label{table: actions}
\end{table}

\begin{figure}[h]
\begin{center}
\includegraphics[width=0.7\linewidth]{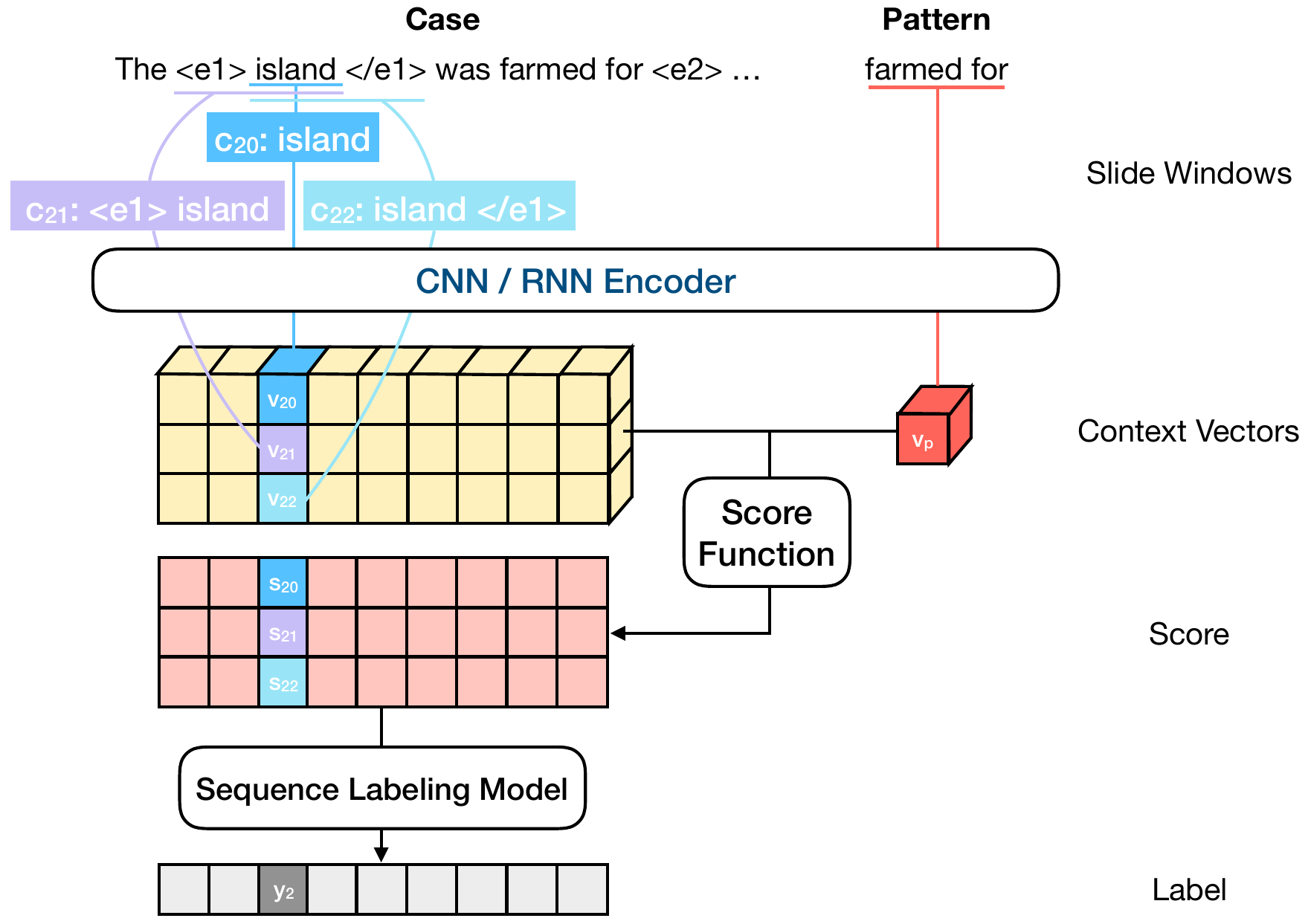}
\end{center}
\caption{``Find'' module. 
Given a case and a pattern, ``Find'' module is designed to label every token of the case, indicating if the token is ``found'' by the pattern. 
The figure shows how to label $y_{2}$ for token $island$. }
\label{fig: find}
\end{figure}

We define 6 basic actions and the most of REs can be interpreted by them. 
As shown in Table \ref{table: actions}, ``Find\_Positive'', ``Find\_Negative'' and ``And\_Ordered'' modules are based on neural networks while the other modules are based on mathematical methods.
``Find'' modules aim to find the related words in a sentence, and then label them. 
The process is similar to sequence labeling.
Since ``Find'' modules may have different preferences (recall and precision) to the positive patterns and the negative patterns, we split ``Find'' modules into ``Find\_Positive'' and ``Find\_Negative''.
``And'' modules (``And\_Ordered'' and ``And\_Unordered'') are designed to process the relation between two groups of labels and output new labels.

``Find'' module is illustrated in Figure \ref{fig: find}.
Given a case $x = [x_{0}, \cdots, x_{n}]$ and an action $Find(p)$, where $p$ is a pattern, ``Find'' Module aims to label 0 or 1 (1 indicates this token is matched with $p$ ) for every token of $x$.
Different contexts are obtained by slide windows of various lengths and each of them is encoded to a fixed-length vectors by a neural network:
\begin{sequation}
\label{for: encoder}
v_{ij} = NN(c_{ij})
\end{sequation}
where $c_{ij}$ is the $j^{th}$ context of the $i^{th}$ word, and $v_{ij}$ is the representation of context $c_{ij}$.
Also, $p$ is encoded to a fixed-length vector $v_{p}$ by the same NN.
Finally, we use Function \ref{for: score} to calculate the scores between every context $c_{ij}$ and fixed-length vector $v_{p}$, and then we decide the label 0 or 1 by the scores using a sequence labeling model, which maps scores $s_i$ to a label $y_i$ at every position of $x$.
\begin{sequation}
\label{for: score}
s_{ij} = Score(v_{ij}, v_p) = v_{ij}^T * W * v_p
\end{sequation}
where W is a matrix that can be trained.
As shown in Figure \ref{fig: find}, given a word $x_2$ and a pattern $p$, we first extract different levels contexts $c = [c_{20}, c_{21}, c_{22}]$ with the window size of 2, where $c_{20}$ is 
$x_{2}$, $c_{21}$ is $x_{1}\textcircled{+}x_{2}$, $c_{22}$ is $x_{2}\textcircled{+}x_{3}$ and $\textcircled{+}$ is the concatenation operator.
Then $c$ and $p$ are encoded to fixed-length vectors $v_{c} = [v_{20}, v_{21}, v_{22}]$ and $v_{p}$ by the same encoder.
After that, we use function \ref{for: score} to calculate scores $s_{2} = [s_{20}, s_{21}, s_{22}]$ between $v_{c}$ and $v_{p}$.
Finally, label $y_{2}$ is output according to $s_{2}$ with a sequence labeling model.

\begin{figure}[h]
\begin{center}
\includegraphics[width=0.7\linewidth]{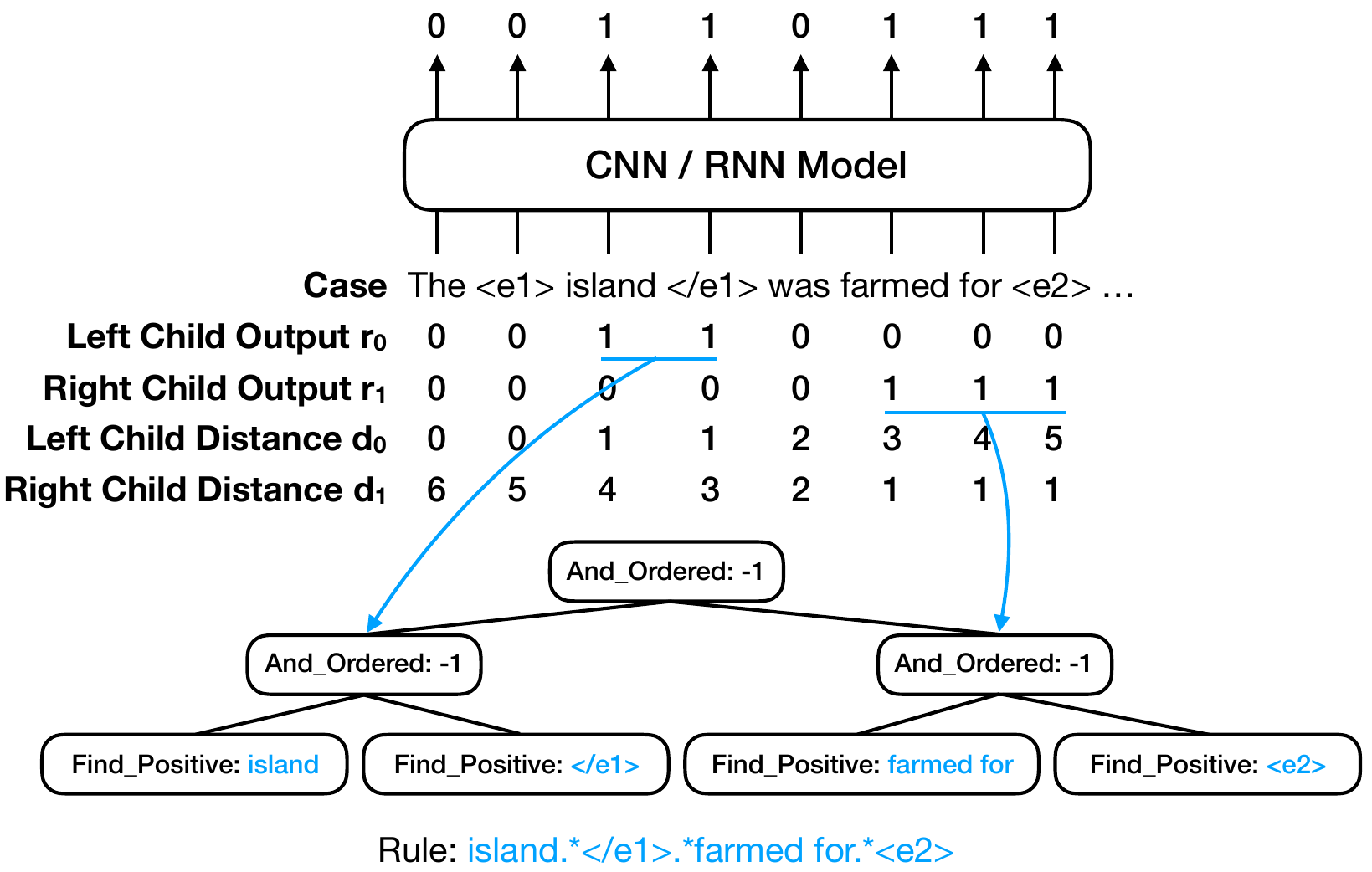}
\end{center}
\caption{``And\_Ordered'' module. 
A rule $The \: \langle e1 \rangle \: island \: \langle \/e1 \rangle \: was \: farmed \: for \: \langle e2 \rangle \cdots$ is parsed to a tree structure as the action sequence. 
The figure shows how the root node ``And\_Ordered'' merges the two label sequences calculated by two subnodes. 
}
\label{fig: AndOrdered}
\end{figure}

As shown in Figure \ref{fig: AndOrdered}, ``And\_Ordered'' module is designed to process the relation between two groups of sequential labels and output new labels. Distance parameter $d$ is required for ``And\_Ordered'' module, indicating the maximum distances between two input sequential labels. Distance $-1$ means that the two input labels can be combined at an arbitrary distance. For the root node ``And\_Ordered'' in Figure \ref{fig: AndOrdered}, the left child output $r_{0}$ is $ 0 \: 0 \: 1 \: 1 \: 0 \: 0 \: 0 \: 0$, indicating that tokens $island$ and $\langle \/e1 \rangle$ are ``found'' respectively by ``Find\_Positive: $island$'' and ``Find\_Positive: $\langle \/e1 \rangle$'', and they are merged by ``And\_Ordered'' in the left node.
The right child output $r_{1}$ is calculated in the same way.
$d_{0}$ and $d_{1}$, indicating the distance from the found token in $r_{0}$ and $r_{1}$, are calculated based on distance $-1$.
\footnote{We conduct experiments with CNN and RNN to determine which model works best. 
Experiments show that CNN and RNN are comparable in accuracy, but CNN has a faster speed.
So we select CNN to implement the modules.}

\subsection{Layout Policy}
\label{sec: Layout Policy}

For every module, it can be a predefined algorithm or a customized neural network. 
Similarly, the rule parser can also be implemented with a predefined algorithm or a neural network.
A neural rule parser can reconstruct rules and make rules more flexible than a deterministic algorithm (Table \ref{table: NN replace algorithm} and Table \ref{table: case of seq2seq}).
As shown in Figure \ref{fig: rule_disassemble}, the layout parser predicts a tree structure for each rule.
We treat the tree structure as a RPN for convenience.
Figure \ref{fig: rpn_example} shows a RE example and its linearized RPN sequence, which consists of actions and parameters.
Then the layout prediction problem turns into a sequence-to-sequence learning problem from REs to modules and their parameters.
We train a novel seq2seq model to predict RPNs from REs.

\subsection{Training Method}
\label{sec: Training Method}

The pretrained fastText \citep{bojanowski2017enriching} word embeddings are used as input, they are kept static during the training.
We will introduce how to train modules and layouts separately.

The training of modules has two phases: pretraining and finetuning.
In the pretraining phase, we generate the training data for modules based on sentences in training set, i.e. we randomly select patterns and sentences that match the patterns.
For a better recall, the maximum epoch number is set to avoid overfitting.
In the finetuning phase, we use an algorithm to transform a rule to a tree.
Then the network is assembled according to the layout and the final prediction is given based on the results of each module. 
We use the final result to finetune the modules with Reinforcement Learning.
Taking ``Find\_Positive'' modules as an example, a ``Find\_Positive'' module is to label 0 or 1 for every token of $x$ according to $p_{i}$. 
If the final prediction contains $y_{i}$ while the ground truth doesn't, we believe that ``Find\_Positive'' modules ``Find'' too many tokens and we punish ``Find\_Positive'' modules whose predictions are 1.

\begin{figure}[h]
\begin{center}
\includegraphics[width=0.7\linewidth]{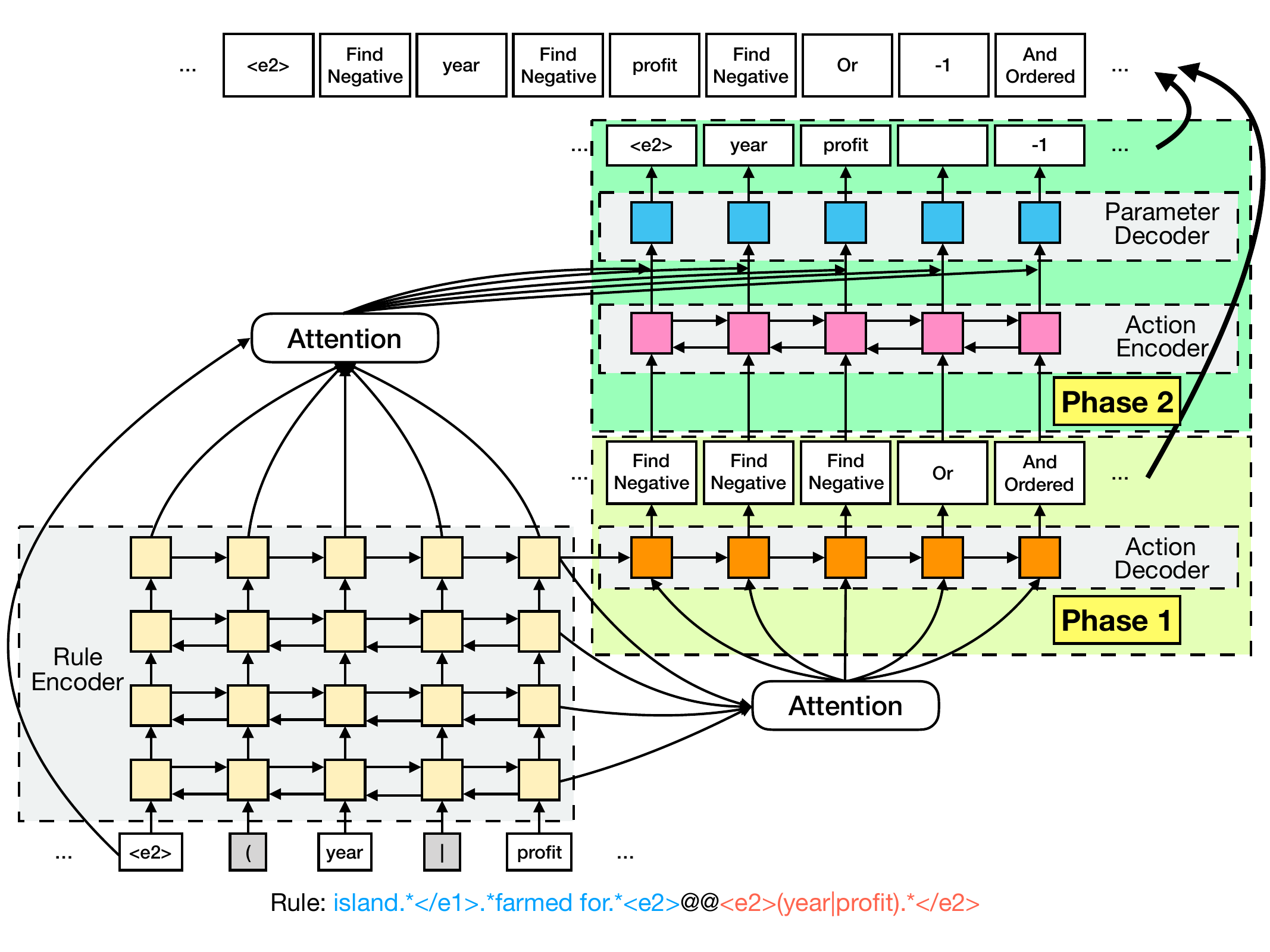}
\end{center}
\caption{An seq2seq model to predict a layout.}
\label{fig: layout_policy}
\end{figure}

The model for layout parser is shown in Figure \ref{fig: layout_policy}.
Given a rule, the layout parser is to predict a specific RPN, consisting of actions and parameters.
Considering it is difficult to predict actions and parameters at the same time, we split the training process into three phases:
\begin{enumerate}[label=\textbf{Phase \arabic*. }, leftmargin=0.7in]
\item{predict actions}
\item{predict parameters based on the actions and the input sentence}
\item{jointly finetune actions and parameters}
\end{enumerate}

In Phase 1, we use an attentional seq2seq model to predict actions. As shown in Figure \ref{fig: layout_policy}, the encoder is a three-layer bidirectional LSTM and a one-layer unidirectional LSTM, while the decoder is a one-layer LSTM.
Beam Search is used to decode actions during prediction.

In Phase 2, to predict parameters for each action, we add an action encoder (a bidirectional LSTM) to the rule encoder of Phase 1.
The parameter decoder is also a one-layer LSTM.
During the training, the parameters of an action need to be searched across the entire vocabulary, which is quite inefficient.
To solve this problem, we propose a novel strategy to change the optimization target from Function \ref{for: target old} to Function \ref{for: target new}, i.e., force the model to predict a word vector rather than to directly find out the id of the target word. 
Since fixed word vectors are used, this strategy can greatly speed up training, and does not affect the accuracy. 
\begin{sequation}\label{for: target old}
loss = \sum_{i=1}^{n} \sum_{j=1}^{m} \hat{y}_{ij} * \log y_{ij} + (1 - \hat{y}_{ij}) * \log (1 - y_{ij})
\end{sequation}
\begin{sequation}\label{for: target new}
loss = \sum_{i=1}^{n} ||vec(\hat{y}_i) - vec(y_{i})||_{2}
\end{sequation}
\hspace{-0.3cm}where $n$ is the length of the sentence, and $m$ is the vocabulary size.
At the time of prediction, we predict a vector and consider the closest word as the predicted word based on all word vectors.

In Phase 3, we adopt the same strategy as in the module training to adjust the layout parser.
According to the predicted layout, we assemble the network with trained modules and get the final result.
Then Reinforcement Learning is exploited to finetune the parser based on the ground truth. 
We encourage any changes in the rule parser if modules and the layout predicted by the parser give a right label as same as the true label of the case. 
The rule parser will be punished if it assigns a wrong label to the case.
Although the true labels are used, it should be noted that NRE only focuses on the matching rather than learning the true label of the case from scratch. 
The true labels are utilized to help NRE generalize rules in the right direction.

\section{Experiment}

\subsection{Baseline Models}

\begin{figure}[t!]
\centering
\begin{subfigure}[t]{\linewidth}
\centering
\includegraphics[width=0.5\linewidth]{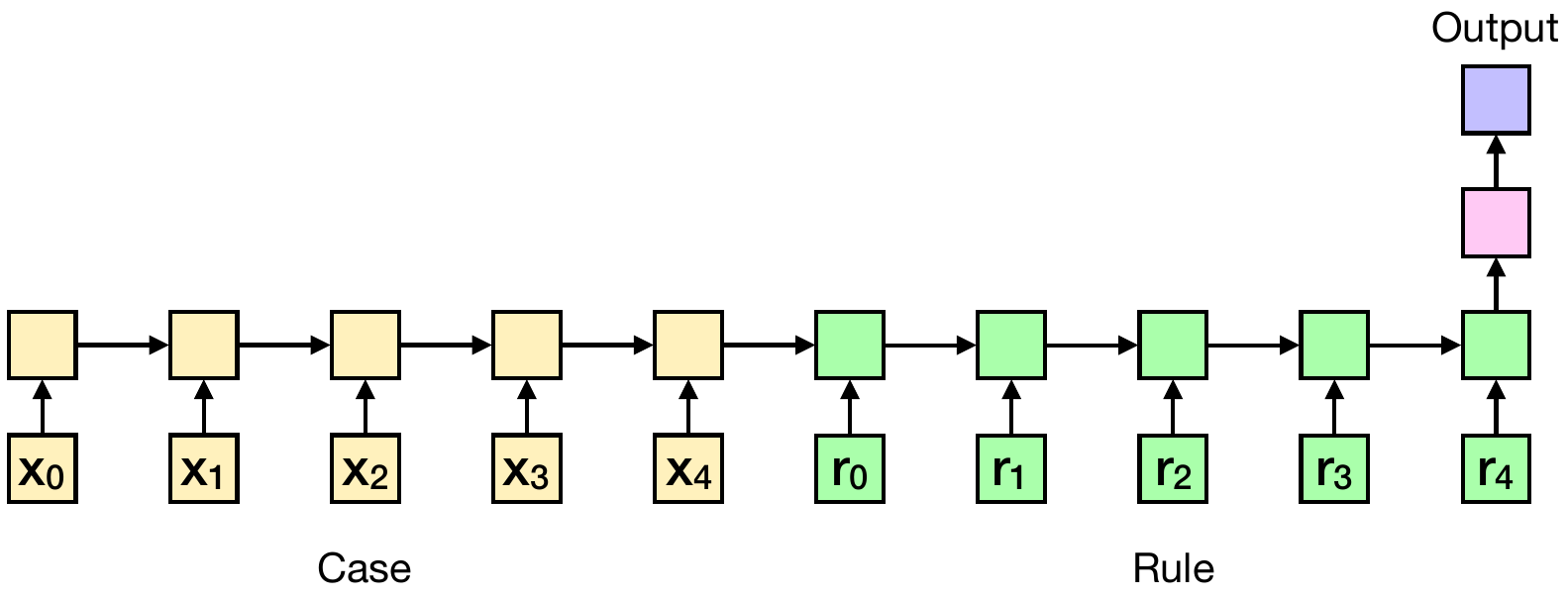}
\subcaption{}
\label{fig:baseline_no_attetion}
\end{subfigure}
\begin{subfigure}[t]{\linewidth}
\centering
\includegraphics[width=0.5\linewidth]{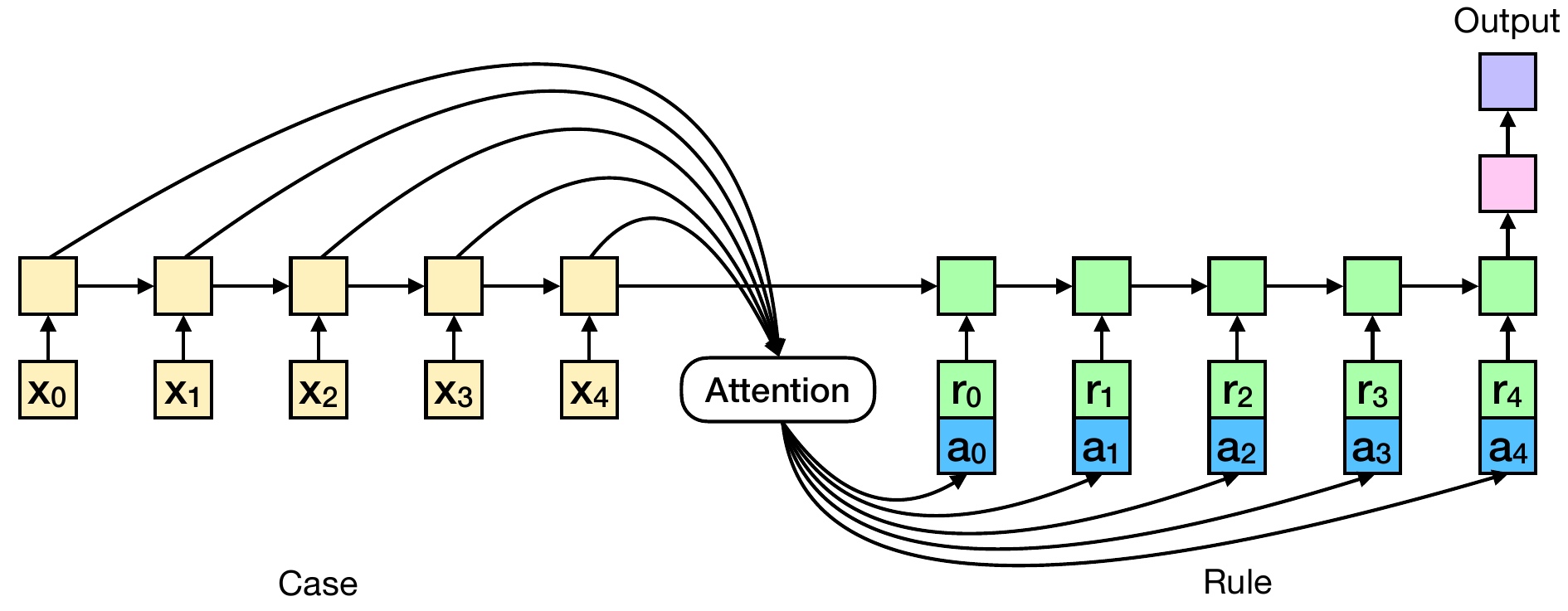}
\subcaption{}
\label{fig:baseline_attention_input}
\end{subfigure}
\begin{subfigure}[t]{\linewidth}
\centering
\includegraphics[width=0.5\linewidth]{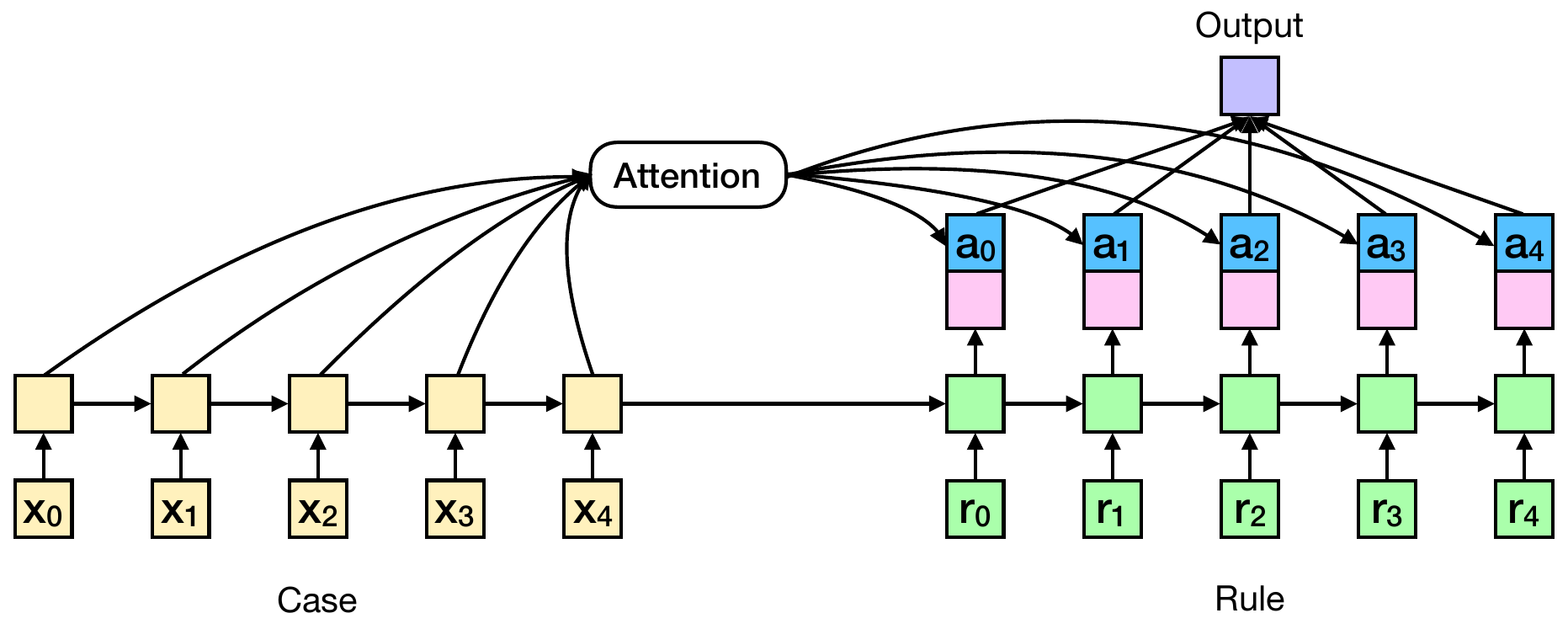}
\subcaption{}
\label{fig:baseline_attention_output}
\end{subfigure}
\caption{Figure \ref{fig:baseline_no_attetion} is the basic sequence model, called ``LSTM-No-Attention''. 
  Figure \ref{fig:baseline_attention_input} and Figure \ref{fig:baseline_attention_output} are attentional sequence models, which are ``LSTM-Attention-Input'' and ``LSTM-Attention-Output''.}
\label{fig:baselines}
\end{figure}

To evaluate the generalization capabilities of NRE, we apply it to classification tasks. 
Two types of baselines are proposed: RE based and NN based.
It should be noted that the task is achieved by matching cases and classification rules like rule systems.
Given a case and a rule set, RE baselines directly output labels of the matched rules, and the NN baselines are trained to learn whether a rule matches the case, and assign the label of the matched rule to the case.
The RE baseline has two methods: traditional RE and the improved ``RE-Synonyms'' which uses synonyms\footnote{\href{https://ltp-cloud.com/download/}{https://ltp-cloud.com/download/}} to generalize the rules. 
NN baselines are implemented with LSTM and attention mechanism, including three variants as shown in Figure \ref{fig:baselines}.

``LSTM-No-Attention'' is illustrated in Figure \ref{fig:baseline_no_attetion}.
Given a RE $r_{y_{m} i} = [r_{1}, \cdots, r_{k}]$, and a case $x = [x_{1}, \cdots, x_{n}]$, the case and the rule are both encoded with LSTM-No-Attention. The final hidden state of the case LSTM is used as the initial hidden state of the rule LSTM.
We use the final state of the rule LSTM to get prediction.
The ``LSTM-Attention-Input'' baseline is inspired by \citep{bahdanau2014neural}. As shown in Figure \ref{fig:baseline_attention_input}, we force the RE to attend different parts of the case at each step of RE input.
The last baseline model ``LSTM-Attention-Output'' is shown in Figure \ref{fig:baseline_attention_output}.
Given the case outputs of all timestamps $h_{c} = [h_{c1}, \cdots, h_{cn}]$ and the RE outputs $h_{r} = [h_{r1}, \cdots, h_{rm}]$, we force the case outputs $h_{c}$ to attend different parts of $h_{r}$ and concatenate them as $\hat{h}_{r} = h_{c}\textcircled{+}attention(h_{r})$.  
Inspired by max-pooling in CNN \citep{collobert2011natural}, which encourages the network to capture the most useful local features after convolutional layers, we utilize max-pooling on $\hat{h}_{r}$ to predict final labels.

\subsection{Data and Hyperparamters}

We use two datasets to evaluate the performance of the proposed approach: the Criminal Case Classification dataset in Chinese \citep{lu2017object} and the relation classification dataset in English \citep{hendrickx2009semeval}\footnote{We released rules on Github:
 \href{https://github.com/}{https://github.com/}.
}.

In Chinese Criminal Case Classification, each case contains one to three sentences recorded by policemen.
There are 150 labels in the dataset, such as ``Burglary'', ``Motorcycling Robbery''.
The entire dataset consists of 12555 cases and each case corresponds to zero, one or more than one labels.
We build 239 rules with RE to label the cases.
A typical example is shown in the top part of Figure \ref{fig: rule example}
Both the dataset and the rule set are split into training, validation and test sets with the ratio of 80\%, 10\%, and 10\%.

Besides, we conduct experiments with the relation classification dataset from ``SemEval-2010 Task 8''.
Each case belongs to one of nineteen relation types.
Among them, there are 7109 cases in training set, 891 in validation set and 2717 in test set.
We construct a rule set, including 365 rules for training, 54 rules for validation and 160 rules for testing.
It should be noted that all the rules are built based on the training dataset.

To train the neural layout parser and the modules, the experiments are conducted through a pretraining phase and a finetuning phase.
It is important that during the pretraining phase, only the training dataset is fed into NRE.
For action modules, patterns are randomly chosen from sentences in the training dataset, by which we can generate massive data.
For the parser, we randomly generate RE-RPN pairs at each training step.
In the finetuning phase, both the training dataset and the training rule set are used by Reinforcement Learning. 
To test the model, we feed the test dataset and the test rule set into NRE which have not been seen by the model.

\begin{table}[!htb]
\begin{center}
\begin{tabular}{cccccc}
\hline
\multicolumn{1}{c}{\bf Module}&\multicolumn{1}{c}{\bf \#Filters} &\multicolumn{1}{c}{\bf Filter Size} &\multicolumn{1}{c}{\bf Slide Window} &\multicolumn{1}{c}{\bf Dropout} &\multicolumn{1}{c}{\bf Embedding Size}\\
\hline
\hline
Find\_Positive &200 &[1, 2, 3] &[1, 2, 3] &0.5 &300\\\hline
Find\_Negative &200 &[1, 2] &[1, 2] &0.5 &300\\\hline
And\_Ordered &100 &[3, 4, 5] &N/A &0.5 &300\\\hline
\end{tabular}
\begin{tabular}{ccccc}
\hline
\multicolumn{1}{c}{\bf Module}  &\multicolumn{1}{c}{\bf RNN Size} &\multicolumn{1}{c}{\bf Beam Width} &\multicolumn{1}{c}{\bf Dropout} &\multicolumn{1}{c}{\bf Embedding Size}\\
\hline
\hline
Rule Parser &500 &3 &0.5 &300\\\hline
\end{tabular}
\end{center}
\caption{Hyperparameters}
\label{table: hyperparamters}
\end{table}

We train models with Adadelta Optimizer \citep{zeiler2012adadelta} and finetune them by Stochastic Gradient Descent (SGD) \citep{kiefer1952stochastic} with learning rate 0.001.
We utilize pretrained fastText \citep{bojanowski2017enriching} word vectors as inputs.
More hyperparameters are shown in Table \ref{table: hyperparamters}.

\subsection{Result}
\label{sec: result}

\begin{figure}
\begin{subfigure}[t]{0.32\linewidth}
\includegraphics[width=\linewidth]{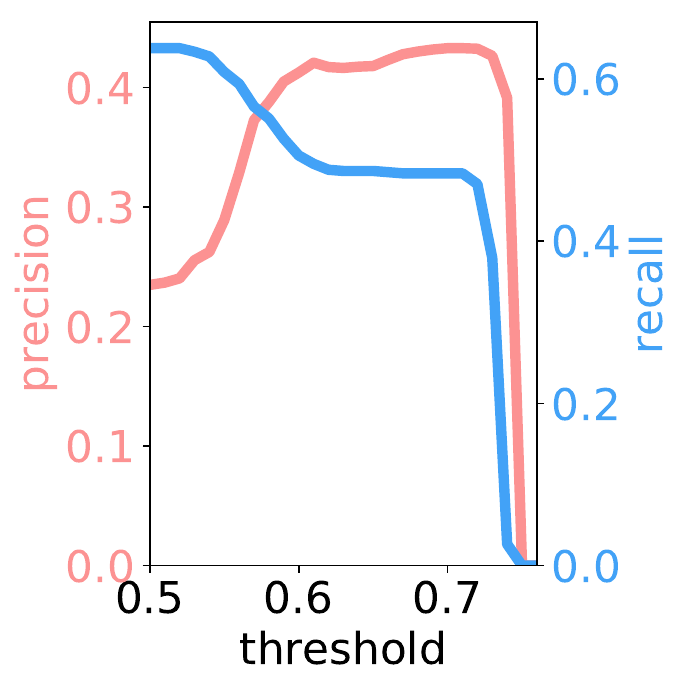}
\subcaption{}
\label{fig:baseline_no_attetion_result}
\end{subfigure}
\begin{subfigure}[t]{0.32\linewidth}
\includegraphics[width=\linewidth]{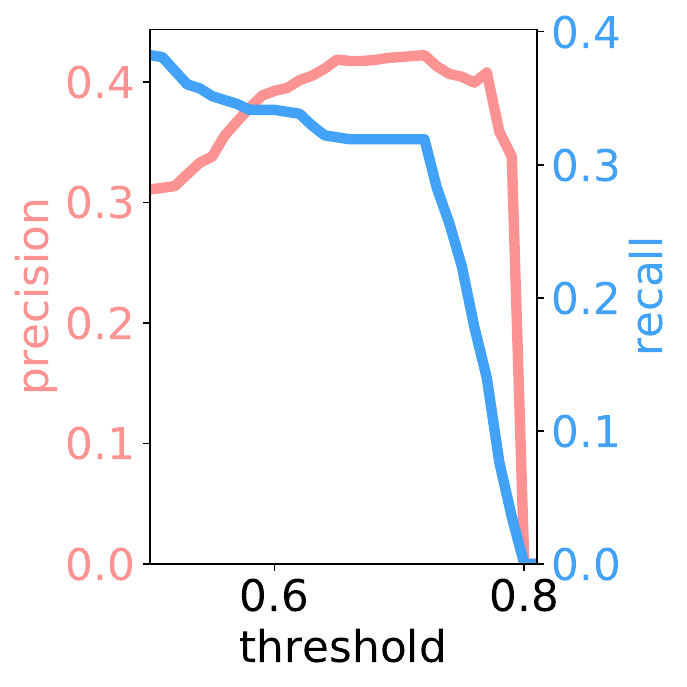}
\subcaption{}
\label{fig:baseline_attention_input_result}
\end{subfigure}
\begin{subfigure}[t]{0.32\linewidth}
\includegraphics[width=\linewidth]{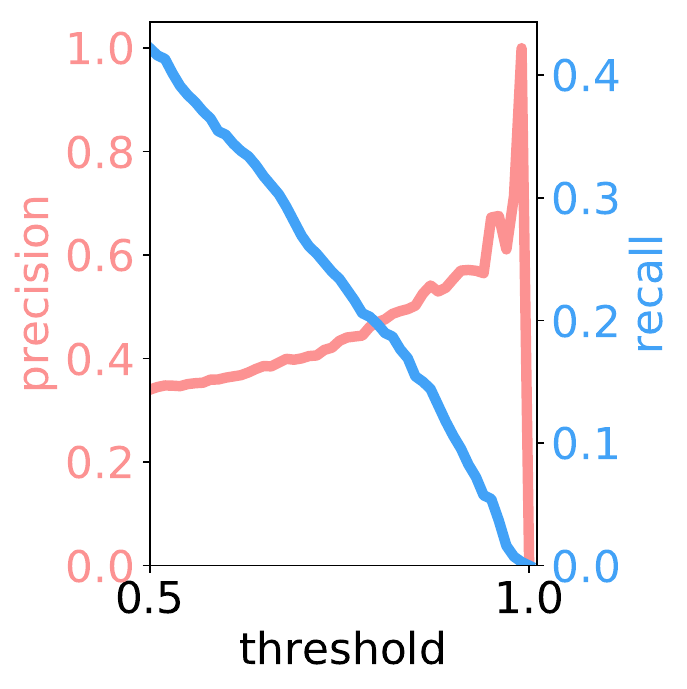}
\subcaption{}
\label{fig:baseline_attention_output_result}
\end{subfigure}
\caption{Tuning the threshold of confidence in baselines. Figure \ref{fig:baseline_no_attetion_result}, Figure \ref{fig:baseline_attention_input_result} and Figure \ref{fig:baseline_attention_output_result} are ``LSTM-No-Attention'', ``LSTM-Attention-Input'' and ``LSTM-Attention-Output'' respectively.}
\label{fig:baselines_result}
\end{figure}

\begin{table}[!htb]
\begin{center}
\begin{tabular}{cccc}
\hline
\multicolumn{1}{c}{\bf Model}  &\multicolumn{1}{c}{\bf Precision} &\multicolumn{1}{c}{\bf Recall} &\multicolumn{1}{c}{\bf F1} \\
\hline
\hline
RE         &1.0 &0.1779 &0.3021\\
RE-Synonyms         &0.9773 &0.1897 &0.3177\\
\hline
LSTM-No-Attention         &0.2348 &0.6382 &0.3434\\
LSTM-Attention-Input         &0.3106 &0.3824 &0.3428\\
LSTM-Attention-Output        &0.3400 &0.4221 &0.3766\\
\hline
NRE-Char         &0.8104 &0.2515 &0.3838\\
NRE-Char-Finetune         &0.9261 &0.2765 &0.4258\\
NRE-Word         &0.6127 &0.3559 &0.4502\\
NRE-Word-Finetune         &0.9237 &0.3382 &0.4952\\
\hline
\end{tabular}
\end{center}
\caption{NRE on the Criminal Case Classification dataset.
With ``Finetune'', NRE increases the recall of rules significantly while maintaining a high precision.}
\label{table: chinese case}
\end{table}

\begin{table}[!htb]
\begin{center}
\begin{tabular}{cccc}
\hline
\multicolumn{1}{c}{\bf Model}  &\multicolumn{1}{c}{\bf Precision} &\multicolumn{1}{c}{\bf Recall} &\multicolumn{1}{c}{\bf F1} \\
\hline
\hline
RE         &0.9143 &0.0589 &0.1107\\
\hline
LSTM-No-Attention         &0.4237 &0.3485 &0.3825\\
LSTM-Attention-Input         &0.5097 &0.1840 &0.2704\\
LSTM-Attention-Output        &0.5983 &0.1266 &0.2090\\
\hline
NRE-Word-Finetune         &0.8837 &0.1121 &0.1990\\
\hline
\end{tabular}
\end{center}
\caption{NRE on the relation classification dataset. }
\label{table: relation classification}
\end{table}

The results of Chinese Criminal Case Classification are summarized in Table \ref{table: chinese case}, where NRE significantly outperforms 2 ``RE'' baselines and 3 LSTM baselines.
Since tokens can be represented as characters (Hanzi) or words in Chinese, we conduct experiments with ``NRE-Char'' and ``NRE-Word'', where cases and rules are based on characters and words respectively.
As illustrated in Section \ref{sec: Training Method},  ``Finetune'' jointly optimizes modules and the layout policy.
We conduct experiments to test whether ``Finetune'' makes effect, and the results are listed as ``NRE-Char-Finetune'' and ``NRE-Word-Finetune''.

It can be seen that ``RE'' achieves 100\% precision and 17.79\% recall, indicating that the rules are accurate but can merely cover a small part of data.
Even though ``RE-Synonyms'' can generalize the rules with synonyms from a dictionary, it just gains a slight improvement in recall.
It is because dictionary-based synonyms only work at the word match level, and they may not be suitable for the dataset.

As shown in Table \ref{table: chinese case}, the LSTM baselines achieve unsatisfactory but reasonable results.
After integrating neural networks, the recall increases to a great extent, but the precision decreases to an unacceptable level.
There is always a trade-off between precision and recall, so we adjust the decision threshold to make futher analysis of the model performances.
As shown in Figure \ref{fig:baselines_result}, the precision is still very low even we adjust the threshold to a very high level. 
The reasons are as follows.
A RE has a hierarchical structure while the LSTM models read it in a linear way.
And special characters in RE such as ``.'', ``?'' are difficult to be modeled by a sequence model.  
Besides, RE is insensitive to the length of context, but a sequence model is effective in capturing local features.
By integrating attention, the model is able to detect long distance dependency between words.
The ``LSTM-Attention-Output'' outperforms ``LSTM-Attention-Input'' since the rule output can directly gain information from the case output, and the most useful pattern is captured by max-pooling.
In brief, the LSTM baselines can not work efficiently in learning the symbolic knowledge from logic rules with hierarchical structures.

Compared to ``RE'', ``NRE-Word-Finetune'' increases the recall of rules significantly while still maintaining a high precision.
It can be seen that ``Finetune'' via Reinforcement Learning is critical since it makes the modules and the layout parser fit to the data.
Moreover, ``NRE-Word-Finetune'' is more effective than ``NRE-Char-Finetune'' because words encode more grammatical and semantic information and have better generalization abilities than characters.

Similarity, Table \ref{table: relation classification} shows the results of relation classification task, where NRE improves the performance of rules by a significant margin.

\subsection{Further Analysis of NRE}

\textbf{Where the generalization comes from.}
We conduct experiments by making different combinations of neural networks and RE algorithms.
The results are shown in Table \ref{table: NN replace algorithm}.
Specifically, ``RE'' can be considered a special NRE model where all modules and the layout parser are implemented by RE algorithms.
``PNAS\_'' is the model with best flexibility and generalization ability where all modules are neural networks.
The recall is improved by a significant margin when we use neural ``Find\_Positive'' and  ``Find\_Negative'' modules to replace the RE algorithm.
The performance is increased again if the ``And\_Ordered'' is replaced by a neural network. 
And it is further improved when the neural layout parser is introduced to replace the predefined algorithm.
Intuitively, ``Find'' modules aim to match the related words with patterns and neural ``Find'' modules are capable of finding words which are not strictly same as the words listed in the rules.
For example, ``Find'' module can find ``put inside'' given ``pushed into''.

\textbf{What the generalization is.}
By analyzing the classification cases, we find that the generalization ability of NRE can be seen from two aspects.
Firstly, NRE is able to match semantically and grammatically similar words.
As shown in Table \ref{table: case in rc}, given a rule including the pattern $extracted \:from$, NRE can match sentences with $taken \: from$.
Given a pattern $originate$, ``Find'' module can find $originating$.
Likewise, $put \:inside$ and $sends \:into$ can be matched when $pushed \:into$ is given in a rule.
Table \ref{table: case in cccc} shows some examples in Chinese.
Secondly, action sequences can also be optimized with the neural layout parser.
Taking the first case in Table \ref{table: case of seq2seq} as an example, some actions and parameters are removed, which leads to better performance, such as ``落水管'' (downpipe) and ``不锈钢管'' (stainless steel pipe) are refined to ``管'' (pipe). 

\begin{table}[!htb]
\begin{center}
\begin{tabular}{cccc}
\hline
\multicolumn{1}{c}{\bf Model}  &\multicolumn{1}{c}{\bf Precision} &\multicolumn{1}{c}{\bf Recall} &\multicolumn{1}{c}{\bf F1} \\
\hline
\hline
RE         &1.0 &0.1779 &0.3021\\
\hline
PN\_AS         &0.9313 &0.3191 &0.4754\\
PNA\_S         &0.9331 &0.3279 &0.4853\\
PNAS\_         &0.9237 &0.3382 &0.4952\\
\hline
\end{tabular}
\end{center}
\caption
{Results on different combinations of neural networks and RE algorithms in NRE.
``P'', ``N'' and ``A'' indicate ``Find\_Positive'', ``Find\_Negative'' and ``And\_Ordered'' modules respectively and ``S'' indicates the action sequence parser.
The parts of NRE implemented by NNs are on the left side of ``\_'' and the parts implemented by RE algorithms are on the right side.
}
\label{table: NN replace algorithm}
\end{table}

\begin{table}[!htb]
\begin{center}
\begin{tabular}{cc}
    \hline
    Label & Entity-Origin \\
    \hline
    RE & \</e1\>.* extracted from .*\< e2\> @@ \\
    \hline
    Case & \specialcell{The \< e1\> woman \</e1\> was {\color{red}taken from} her native\\ \< e2\> family \</e2\>
              and adopted in England in the 1960's.} \\
    \hline
    \hline
    Label & Entity-Origin \\
    \hline
    RE & \</e1\>.* originate .*\</e2\>@@(signal\textbar aberration\textbar reduction).*\</e1\> \\
    \hline
    Case & \specialcell{A \<e1\> nunt \</e1\> is a pastry {\color{red}originating} from Jewish\\ \<e2\> cuisine
              \</e2\> and vaguely resembles nougat.} \\
    \hline
    \hline
    Label & Entity-Destination \\
    \hline
    RE & \</e1\>.* pushed into .*\<e2\>@@\<e2\>(function\textbar care).*\</e2\> \\
    \hline
    Case & \specialcell{Then, the target PET \<e1\> bottle \</e1\> was {\color{red}put inside} of \\a metal
              \<e2\> container \</e2\> , which was grounded.} \\
    \hline
    \hline
    Label & Entity-Destination \\
    \hline
    RE & \</e1\>.* pushed into .*\<e2\>@@\<e2\>(function\textbar care).*\</e2\> \\
    \hline
    Case & \specialcell{NASA Kepler mission {\color{red}sends} \<e1\> names \</e1\> {\color{red}into}\\ \<e2\> space \</e2\>.} \\
    \hline
    \hline
    Label & Entity-Destination \\
    \hline
    RE & \</e1\>.* derived from .*\<e2\>@@ \\
    \hline
    Case & \specialcell{A tidal wave of \<e1\> talent \</e1\> has {\color{red}emanated from} \\this lush
                        \<e2\> village \</e2\>.} \\
    \hline
\end{tabular}
\end{center}
\caption{Examples in Relation Classification.}
\label{table: case in rc}
\end{table}

\begin{table}[!htb]
\begin{center}
\begin{tabular}{cc}
    \hline
    Label & 入室作案 (Burglary) \\
    \hline
    RE & 入室@@死亡\textbar工地 \\
    \hline
    Case & \specialcell{2008年9月4日2时许，武进县霞榕路4号发生一起盗窃案。受\\
              害人籍希逵称，小偷攀爬阳台打开未锁的窗，{\color{red}进入室内}，\\
              将其放在房内的手机盗走，损失物品：手机一部诺基亚牌，\\
              型号6030，串号不详，价值500元。嫌疑人不详。} \\
    \hline
    \hline
    Label & 持锐器 (Holding a sharp instrument) \\
    \hline
    RE & 水果刀@@刺绣厂\textbar铁管\textbar马刀帮\textbar被盗\textbar划分\textbar划破 \\
    \hline
    Case & \specialcell{2005年12月8日晨，接110指挥中心指令：城关镇岙桥里62号\\
              有人{\color{red}持刀}抢劫，受害人为倪霞，犯罪嫌疑人逃跑时被抓。} \\
    \hline
    \hline
    Label & 持枪 (Holding a gun) \\
    \hline
    RE & 气枪@@ \\
    \hline
    Case & \specialcell{2008年1月22日2时许，案犯窜至新泰市育才路29号门前，将\\
                        事主高艳的狗用{\color{red}猎枪}打死后盗走，驾灰色小面包车逃走。} \\
    \hline
\end{tabular}
\end{center}
\caption{Examples in Criminal Case Classification.}
\label{table: case in cccc}
\end{table}

\begin{table}[!htb]
\begin{center}
\begin{tabular}{cc}  
    \hline
    Label & 持钝器 (Holding the blunt) \\
    \hline
    RE & 锤子@@围墙\textbar被盗\textbar落水管\textbar钢管厂\textbar不锈钢管 \\
    \hline
    RE Parser & \hspace{-0.2cm} \specialcell{锤子 Find\_Positive 围墙 Find\_Negative 被盗 Find\_Negative Or \\
                {\color{red}落水管} Find\_Negative 钢管厂 Find\_Negative Or Or {\color{red}不锈钢} \\
                Find\_Negative {\color{red}管} Find\_Negative 0 And\_Ordered Or \\
                And\_Unordered Output} \\
    \hline
    Neural Parser & \hspace{-0.2cm} \specialcell{锤子 Find\_Positive 围墙 Find\_Negative 被盗 Find\_Negative \\
                Or 钢管厂 Find\_Negative Or {\color{red}管} Find\_Negative Or Or Output} \\
    \hline
    \hline
    Label & 墙上挖洞 (Digging holes in the wall) \\
    \hline
    RE & 打墙洞@@ \\
    \hline
    RE Parser & \hspace{-0.2cm} \specialcell{打 Find\_Positive 墙 Find\_Positive 0 And\_Ordered {\color{red}洞} \\Find\_Positive 
                0 And\_Ordered Output} \\
    \hline
    Neural Parser & \hspace{-0.2cm} \specialcell{打 Find\_Positive 墙 Find\_Positive 0 And\_Ordered Output} \\
    \hline
\end{tabular}
\end{center}
\caption{Action sequences in Criminal Case Classification.}
\label{table: case of seq2seq}
\end{table}

\section{Conclusion}

In this paper, we present a novel learning strategy where the neural networks and symbolic knowledge are combined from the knowledge-driven side.
Based on this learning strategy, we propose Neural Rule Engine (NRE), where rules obtain the flexibility and generalization ability of neural networks, while still maintaining the high precision and interpretability.
NRE is able to learn knowledge explicitly from logic rules and generalize them implicitly with neural networks.
NRE consists of action modules and a rule parser, both of which can either be customized neural networks or a symbolic algorithm.
Given a rule, NRE first predicts a specific layout and then modules are dynamically assembled by the layout to output the result.
Besides, a staged training method is proposed where we first pretrain modules and the neural rule parser, and then use Reinforcement Learning to jointly finetune them.
NRE is not only an innovative paradigm of neural-symbolic learning, but also an effective solution to industrial applications, e.g. upgrading the existing rule-based systems and developing neural rule approaches which do not rely on a mass of training data.

\end{CJK}

\bibliography{iclr2019_conference}
\bibliographystyle{iclr2019_conference}

\end{document}